\documentclass[letterpaper, 10 pt, conference]{ieeeconf} 

\IEEEoverridecommandlockouts                             

\usepackage{graphicx}
\usepackage{caption}
\usepackage{booktabs}
\usepackage{xcolor}
\usepackage{cite}
\usepackage{subcaption}

\title{\LARGE \bf Open-Vocabulary Part-Based Grasping}

\author{Tjeard van Oort, Dimity Miller, Will N. Browne, Nicol\'as Marticorena, Jesse Haviland, Niko Suenderhauf
\thanks{T.O, D.M, W.B, N.M, J.H, N.S acknowledge continued support from the Queensland University of Technology (QUT) through the Centre for Robotics. Furthermore, we wish to acknowledge the support of the Research Engineering Facility (REF) team at QUT for the provision of expertise and research infrastructure in enablement of this project.}}

\begin{document}

\maketitle
\thispagestyle{empty}
\pagestyle{empty}

\begin{abstract}
Many robotic tasks require grasping objects at specific object parts instead of arbitrarily, a crucial capability for interactions beyond simple pick-and-place, such as human-robot interaction, handovers, or tool use. Prior work has focused either on generic grasp prediction or task-conditioned grasping, but not on directly targeting object parts in an open-vocabulary way. We propose AnyPart, a modular framework that unifies open-vocabulary object detection, part segmentation, and 6-DoF grasp prediction to enable robots to grasp user-specified parts of arbitrary objects based on natural language prompts. We evaluate 16 model combinations, and demonstrate that the best-performing combination achieves 60.8\% grasp success in cluttered real-world scenes at 60$\times$ faster inference than existing approaches. To support this study, we introduce a new dataset for part-based grasping and conduct a detailed failure analysis. Our core insight is that modularly combining existing foundation models unlocks surprisingly strong and efficient capabilities for open-vocabulary part-based grasping without requiring additional training.

\end{abstract}

\section{INTRODUCTION}

Robots in human environments must grasp objects in semantically meaningful ways, targeting specific parts rather than arbitrary contact points. For instance, when asked to ``bring a cup of coffee'', a robot should grasp the mug by the handle, making it easy to hand over. If instead asked to ``cut vegetables'', it must grasp the knife by the handle, not the blade. These examples highlight a fundamental capability for interactive and safe manipulation: part-specific grasping based on user intent. A robot should be capable of grasping \emph{any object} by \emph{any part} in an open-vocabulary manner.

While recent grasping systems produce robust 6-DoF predictions~\cite{kleeberger2020single, song2020grasping, sundermeyer2021contact, gou2021rgb}, they lack the semantic understanding to produce grasps at specific object parts. Task-oriented grasping method~\cite{detry2017task, song2015task, murali2021same} incorporate higher-level goals, but these systems do not allow users to specify object parts explicitly, nor do they generalise to arbitrary language prompts.

Lerf-TOGO~\cite{rashid2023language} recently introduced open-vocabulary part-based grasping by combining language-driven radiance fields with grasp synthesis. 
While effective, Lerf-TOGO requires over 100 seconds to train a language-embedded radiance field from multiple views so it can generate a single grasp. This makes it impractical for interactive or real-time use.

\begin{figure}
    \centering
    \includegraphics[width=1\linewidth]{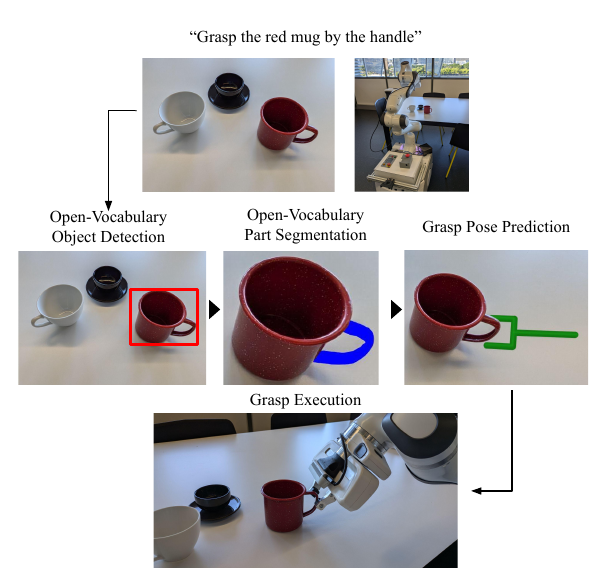}
    \caption{
    AnyPart enables open-vocabulary part-based grasping. Given a natural language query such as ``grasp the red mug by the handle'', AnyPart uses foundation models to detect the relevant object, segment the referred part, and generate 6DoF grasps. AnyPart demonstrates how open-vocabulary part understanding enables targeted interaction with everyday items beyond closed training vocabularies.}
    \label{fig:hero}
\end{figure}

In this paper, we introduce \textbf{AnyPart}, a modular framework for open-vocabulary part-based grasping that produces grasp proposals in less than two seconds. 
AnyPart decomposes the task into three stages: object detection, part segmentation, and grasp prediction, and integrates state-of-the-art foundation models in each module. This modularity lets AnyPart flexibly leverage new foundation models as they improve.

We conduct extensive experiments on 1,000 real-world grasping trials using a Franka Emika Panda arm, and release two new datasets: AnyPart-Vision, a labelled benchmark for object-part segmentation, and AnyPart-Physical, a dataset of physical grasp attempts with success labels. Our results show that AnyPart achieves 60.8\% success in cluttered scenes, close to Lerf-TOGO’s 69\%, while being up to 100$\times$ faster. 

Our experiments yield three key insights. First, existing foundation models can be composed without retraining to enable effective open-vocabulary part-based grasping. Second, the resulting system operates in 1–2 seconds,
making real-time semantic grasping feasible. Third, our failure analysis reveals that performance is bottlenecked by semantic perception, particularly part segmentation and object detection, rather than grasp prediction. This highlights vision-language segmentation as a critical path for future improvement.

In summary, our contributions are:
\begin{enumerate}
    \item AnyPart: A modular framework for fast open-vocabulary part-based grasping.

    \item An empirical evaluation of 16 combinations of object detectors, part segmenters, and grasping models.

    \item Two datasets: AnyPart-Vision for part segmentation benchmarking, and AnyPart-Physical for real-world grasping evaluation.

    \item A detailed failure analysis revealing current limitations in vision-language models for manipulation.
\end{enumerate}

\section{Related Works}

\textbf{Part-Based Grasping: }
Part-aware grasping aims to enable semantically and functionally meaningful interaction with objects. Early work~\cite{aleotti2011part, vahrenkamp2016part, murali2021same, tang2023task, tang2023graspgpt, tang2025foundationgrasp} relied on custom datasets with annotated object parts, which limits scalability and generalisation. While datasets such as ContactDB~\cite{brahmbhatt2019contactdb}, CAGE~\cite{liu2020cage}, and PartNet~\cite{fang2020learning} provide structured part annotations, they are expensive to expand and often constrained to a fixed object taxonomy.

Lan-Grasp~\cite{mirjalili2023lan} integrates GPT-4~\cite{achiam2023gpt} and OWL-ViT~\cite{minderer2022simple} to generate grasp proposals conditioned on language-specified object parts. Although modularity was briefly explored in ablations, the method was not evaluated on physical grasping tasks, limiting insight into real-world effectiveness.

AffordGrasp~\cite{du2025finegrasprobustgraspingdelicate} integrates GPT-4o~\cite{hurst2024gpt} and a visual affordance grounding module to predict object and part affordances based on a task prompt, then using AnyGrasp~\cite{fang2023anygrasp} to generate candidate grasp poses in cluttered scenes. While showing capable grasping success rates the method is limited in only being able to grasp parts that GPT-4o predicts are acceptable.

FineGrasp~\cite{tang2025affordgraspincontextaffordancereasoning}, based the work of EconomicGrasp~\cite{wu2024economicframework6dofgrasp}, implement improvements to the grasp prediction module, paired with Sa2VA-4B~\cite{yuan2025sa2vamarryingsam2llava} for object and part grounding, to perform delicate object grasping. While evaluated on real-world part-grounding tasks, their evaluation covered few objects with one part each, limiting understanding of how their method performs on a larger scale.

We do not extend our evaluation to include AffordGrasp \cite{du2025finegrasprobustgraspingdelicate} and FineGrasp \cite{tang2025affordgraspincontextaffordancereasoning} as both do not have publicly available code for implementing their approaches on real-world scenarios.

Lerf-TOGO~\cite{rashid2023language} uses language-embedded radiance fields to produce 6-DoF grasp proposals for object parts such as “can opener, handle.” Despite strong part grounding, it suffers from slow inference times (upwards of 118 seconds per scene), making it unsuitable for real-time deployment.

\textbf{Open-Vocabulary Perception: }
Open-vocabulary object detection enables flexible language-based interaction in unstructured settings. Detectors such as GroundingDINO~\cite{liu2024grounding}, OWL-ViT~\cite{minderer2022simple}, OWLv2~\cite{minderer2023scaling}, and OmDet-Turbo~\cite{zhao2024real} allow arbitrary object queries via natural language prompts, removing the need for retraining when novel objects are encountered.

Segmentation remains more challenging, particularly at the part level. While general-purpose open-vocabulary segmentation models exist e.g., SAM~\cite{kirillov2023segment}, SEEM~\cite{zou2023segment}, Grounded-SAM~\cite{ren2024grounded}, and FreeSeg~\cite{zhang2023faster}, they focus on object-level masks and are not tuned for fine-grained part segmentation. At the time of writing, VLPart~\cite{sun2023going} is the only known open-vocabulary part segmentation model with public weights, offering a foundation for language-driven part identification.

\textbf{Grasp Pose Prediction: }
Grasp pose prediction methods range from top-down approaches to fully 6-DoF pipelines. Traditional methods~\cite{johns2016deep, satish2019policy, zeng2022robotic} often rely on top-down cameras and bin-picking scenarios, with reduced performance under arbitrary viewpoints. Learned models such as Contact-GraspNet~\cite{sundermeyer2021contact} and AnyGrasp~\cite{fang2023anygrasp} leverage RGB-D or point clouds to predict 6-DoF grasps across diverse scenes. However, these methods typically operate on whole-object inputs and do not incorporate semantic part-level conditioning into their grasp predictions.

\textbf{Summary: }
Together, these works establish the components needed for language-informed robotic grasping: open-vocabulary detection, part-aware segmentation, and 6-DoF grasp prediction. However, prior approaches tend to be constrained by fixed vocabularies, limited part-level resolution, or long inference times. The integration of these components into a modular, efficient, and part-specific grasping system remains an open challenge in robotic manipulation.

\section{Method}
\subsection{Task Definition and Assumptions}

We address the task of open-vocabulary part-based grasping: Given an RGB-D observation and a natural language prompt specifying both an object and one of its parts (e.g., ``mug, handle''), the system must predict a 6-DoF grasp pose targeting that part. A grasp is successful if the robot stably lifts the target object by the target part.

We assume a robot with a wrist-mounted RGB-D camera and a parallel-jaw gripper. The specified object and part must be visible and reachable. In scenes with multiple similar objects, the language prompt should include disambiguating attributes such as color (e.g., ``red mug'').

\subsection{AnyPart Framework}
Our AnyPart framework decomposes open-vocabulary part-based grasping into three sequential stages:  Object Detection, Part Segmentation, and 6-DoF Grasp Prediction. Each stage uses a plug-and-play model and passes its output to the next, as shown in Fig.~\ref{fig:hero}. The resulting overall system takes as input a natural language prompt and an RGB-D image and outputs a grasp pose targeting the specified part.

\textbf{Object Detection: }
The first stage identifies and localises the target object. We use an open-vocabulary object detector that takes an RGB image and an object prompt (e.g. ``red mug'') and returns bounding boxes for candidate detections. We rank these by confidence and select the top box.

To improve downstream segmentation robustness, we expand the selected bounding box's scale by 5\% and crop the image accordingly. This cropped image is passed to the next stage.

We evaluate a total of eight variants of four open-vocabulary detectors of varying size, speed, and architecture:
\begin{itemize}
    \item GroundingDINO (Tiny, Base)~\cite{liu2024grounding}
    \item OWL-VIT-B/32, OWL-VIT-L/16~\cite{minderer2022simple}
    \item OWLv2-B/16 (Base, Fine-tuned, Ensemble)~\cite{minderer2023scaling}
    \item OmDet Turbo~\cite{zhao2024real}
\end{itemize}

\textbf{Part Segmentation: }
The second stage segments the target part from the cropped image. Given the cropped object image and a part prompt (e.g. ``handle''), we use an open-vocabulary part segmenter to produce a binary mask of the part.
This mask is upsampled to the original image resolution and aligned with the depth map for grasp prediction.

We use VLPart~\cite{sun2023going}, the only publicly available open-vocabulary part segmentation model at the time of writing, and evaluate it with two backbones: 
\begin{itemize}
    \item VLPart with a ResNet-50 backbone
    \item VLPart with a SwinBase backbone
\end{itemize}

\textbf{6-DoF Grasp Prediction: }
The final stage predicts a grasp pose targeting the segmented part. It receives the RGB-D image and binary segmentation mask of the target part as input and outputs a ranked list of 6-DoF grasp poses.

We evaluate two grasp prediction models: 
\begin{itemize} 
    \item Contact-GraspNet~\cite{sundermeyer2021contact}, which natively supports mask-conditioned prediction, i.e. only generates grasps in a given bounding box;
    \item AnyGrasp~\cite{fang2023anygrasp}, which requires post-filtering
\end{itemize}
For AnyGrasp, we filter predicted grasps by reprojecting them into the camera view and retaining only those overlapping with the segmented part.

\section{AnyPart-Vision Dataset}
\begin{figure*}
    \centering
    \includegraphics[width=1\linewidth]{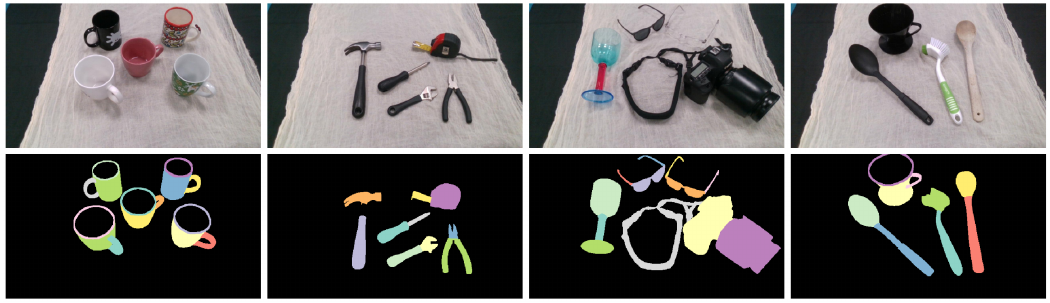}
    \caption{A sample of images taken from the AnyPart-Vision Dataset. In order from left to right are photos of item groups: Mugs, Tools, Fragile, and Kitchen Utensils with the RGB images on top and their hand drawn segmentations on the bottom.}
    \label{fig:Dataset Images}
\end{figure*}

\begin{table}[tb]
    \centering
    \caption{Objects and their parts in the AnyPart-Vision dataset. Most objects have two or more graspable parts.}
    \label{tab:dataset objects-parts}
    \setlength{\tabcolsep}{4.5pt}
    \begin{tabular}{lllll} 
    \toprule
    \textbf{Object} & \textbf{Part 1} & \textbf{Part 2} & \textbf{Part 3} & \textbf{Part 4} \\
    \midrule
    soap bottle & body & cap & - & - \\
    spray bottle & body & cap & - & - \\
    sriracha bottle & body & cap & -  & - \\
    water bottle & body & cap & - & - \\
    corkscrew & handle & screw & - & - \\
    measuring cup & body & handle & rim & - \\
    pan & bowl & handle & - & - \\
    card reader & body & cable & - & - \\
    power board & body & cable & plug & -  \\
    computer mouse & body & cable & - & - \\
    e-stop & box & button & thick cable & thin cable \\
    flashlight & body & button & - & - \\
    camera & body & lens & strap & - \\
    colored wine glass & base & head & stem & - \\
    glasses & left frame & right frame & lens frame & - \\
    black matte spoon & handle & head & - & - \\
    coffee strainer & base & body & handle & rim \\
    scrub brush & handle & head & - & - \\
    wooden spoon & handle & head & - & - \\
    box cutter & handle & - & - & - \\
    bread knife & blade & handle & - & - \\
    kitchen knife & blade & handle & - & - \\
    black mug & body & handle & rim & - \\
    colorful mug & body & handle & rim & - \\
    green mug & body & handle & rim & - \\
    pink mug & body & handle & rim & - \\
    white mug & body & handle & rim & - \\
    hammer & handle & head & - & - \\
    pliers & handles & head & - & - \\
    screwdriver & handle & shaft & - & - \\
    tape measure & body & tape & - & - \\
    wrench & head & handle & - & - \\
    \bottomrule
    \end{tabular}%
\end{table}

To evaluate the perception components of the AnyPart pipeline (object detection and part segmentation) we introduce AnyPart-Vision, a hand-labelled dataset of object-part annotations designed for robotic manipulation scenarios. Sample images and segmentations from AnyPart-Vision are shown in Figure~\ref{fig:Dataset Images}.

\textbf{Motivation: }
Existing part segmentation datasets such as PACO~\cite{ramanathan2023paco} or PascalPart~\cite{hoiem2009pascal}  include many objects irrelevant to robotic manipulation (e.g., cars, ladders), and often annotate parts that are not graspable (e.g., text, logos). These datasets are not well-suited for evaluating part-based grasping systems in realistic tabletop settings.

To address this, we construct a dataset specifically focused on common household and office items that are feasible for manipulation by a robot arm, with parts that are functionally relevant for grasping.

\textbf{Dataset Composition: }
The AnyPart-Vision dataset includes 32 unique objects across common categories such as bottles and containers (e.g. spray bottles, mugs), kitchen tools (e.g. knives, spoons, pans), electronics and cables (e.g. power boards, computer mice) and hand tools (e.g. screwdrivers, hammers). Each object is annotated with 2–4 graspable parts, for a total of 77 hand-segmented part masks. A full list of objects and parts is provided in Table I.

\textbf{Data Collection and Augmentation: }
For each object, we collect an image from the perspective of a wrist-mounted RGB camera, simulating the viewpoint during a real grasp. Binary segmentation masks are manually drawn for each part. Each image is also paired with the corresponding object and part text prompts used by the models during evaluation.

To increase diversity and robustness, we augment the dataset by applying horizontal flips and rotations from -90$^\circ$ to +90$^\circ$ in 30$^\circ$ increments.
This results in a total of 2,212 images, each paired with prompts and part masks.

\textbf{Purpose and Usage: }
This dataset is used to systematically evaluate the first two stages of the AnyPart pipeline: object detection and part segmentation. It enables us to test different combinations of detectors and segmenters in a controlled, single-image setting, without the additional variables introduced by grasp proposal and execution.

\section{Experiments}

We evaluate AnyPart in two stages: first, the perception components (object detection and part segmentation), and second, the full pipeline including grasp prediction in real-world robotic trials.

These experiments support three key claims:

\begin{itemize}
    \item \textbf{Claim 1:} Composing existing foundation models without retraining enables effective open-vocabulary part-based grasping.
    \item \textbf{Claim 2:} AnyPart operates in near real-time (1-2s), significantly faster than prior work while achieving competitive performance.
    \item \textbf{Claim 3:} The primary bottleneck lies in perception (detection and segmentation), not grasp prediction.
\end{itemize}

\begin{table*}[tb]
    \centering
    \caption{Results of the AnyPart-vision evaluation. 
    Numbers in \textbf{bold} indicate the best performance.}
    \resizebox{\textwidth}{!}{
    \begin{tabular}{@{}lcccccccccc@{}}
        \toprule
        Object Detector & VLPart & Scenes with & Scenes with & Object Detector & Object Detector & Part Segmenter & max GPU \\
        & Backbone & one Object  & multiple Objs & avg. IoU  & avg. Inference  & avg. Inference  & Memory \\
        & & mIoU $\uparrow$ & mIoU $\uparrow$ & \textless 0.8 $\downarrow$ & Time (s) $\downarrow$ & Time (s) $\downarrow$ &  (GB) $\downarrow$ \\ 

        \midrule
        Grounding DINO T & ResNet 50 & 0.332 & 0.232 & 0.320 & 0.249 & 1.418 & \textbf{5.875} \\
        Grounding DINO B & ResNet 50 & 0.334 & 0.222 & 0.307 & 0.341 & 1.414 & 6.551 \\
       OWL-VIT-B/32 & ResNet 50 & 0.326 & 0.169 & 0.438 & 0.061 & 1.385 & 6.227 \\
        OWL-VIT-L/16 & ResNet 50 & 0.334 & 0.181 & 0.363 & 0.140 & 1.600 & 12.094 \\
        OmDet Turbo & ResNet 50 & 0.328 & 0.180 & 0.514 & 0.069 & 1.451 & 6.531 \\
        OWLv2-B/16 & ResNet 50 & 0.337 & 0.255 & 0.315 & 0.415 & 1.654 & 16.929 \\
        OWLv2-B/16 Ensemble & ResNet 50 & 0.332 & 0.259 & \textbf{0.274} & 0.419 & 1.655 & 16.854 \\
        OWLv2-B/16 Fine & ResNet 50 & 0.325 & 0.236 & 0.342 & 0.418 & 1.559 & 16.937 \\
        \midrule
        Grounding DINO T & SwinBase & 0.401 & 0.245 & 0.320 & 0.244 & \textbf{0.872} & 6.309 \\
        Grounding DINO B & SwinBase & 0.397 & 0.244 & 0.307 & 0.329 & 0.905 & 6.904 \\
        OWL-VIT-B/32 & SwinBase & 0.382 & 0.186 & 0.438 & \textbf{0.058} & 0.884 & 6.469 \\
        OWL-VIT-L/16 & SwinBase & 0.384 & 0.196 & 0.363 & 0.127 & 1.015 & 12.123 \\
        OmDet Turbo & SwinBase & 0.404 & 0.203 & 0.514 & 0.065 & 0.963 & 7.218 \\
        OWLv2-B/16 & SwinBase & 0.411 & 0.306 & 0.315 & 0.419 & 1.138 & 17.537 \\
        OWLv2-B/16 Ensemble & SwinBase & \textbf{0.412} & \textbf{0.325} & \textbf{0.274} & 0.418 & 1.162 & 17.399 \\
        OWLv2-B/16 Fine & SwinBase & 0.405 & 0.279 & 0.342 & 0.429 & 1.087 & 17.415 \\
        \bottomrule
    \end{tabular}}
    \label{tab:Vision-Results}
\end{table*}

\subsection{Experiment 1: Perception Evaluation on AnyPart-Vision}

We evaluate 16 combinations of object detectors and part segmenters on the AnyPart-Vision dataset to quantify their segmentation accuracy, speed, and compute requirements.

We report the following metrics:
\begin{itemize}
    \item Mean Intersection-over-Union for single-object scenes.
    \item Mean Intersection-over-Union for group (multi-object) scenes.
    \item Object Detector average Intersection-over-Union$<$0.8: Fraction of bounding boxes with Intersection-over-Union $<$ 0.8 vs. ground truth.
    \item Inference Time: Per-image inference time (in seconds), separately for the object detector and part segmenter.
    \item GPU Memory: Peak usage during inference (GB).
\end{itemize}

\textbf{Findings: }
We summarise the results of these experiments in Table~\ref{tab:Vision-Results} and formulate our key findings below.

\textbf{1) VLPart with SwinBase significantly outperforms ResNet50.}  
Across all detectors, VLPart with SwinBase consistently yields higher mean IoU (both in single- and multi-object scenes), confirming it as the stronger backbone for part segmentation. This aligns with the results of prior benchmarks in~\cite{sun2023going}.

\textbf{2) OWLv2-Ensemble + VLPart-SwinBase is the best-performing combination.}  
This pairing achieves the highest segmentation accuracy while maintaining acceptable inference cost, making it our default for real-world evaluation.

\begin{figure*}[t]
    \centering
    \includegraphics[width=0.48\linewidth]{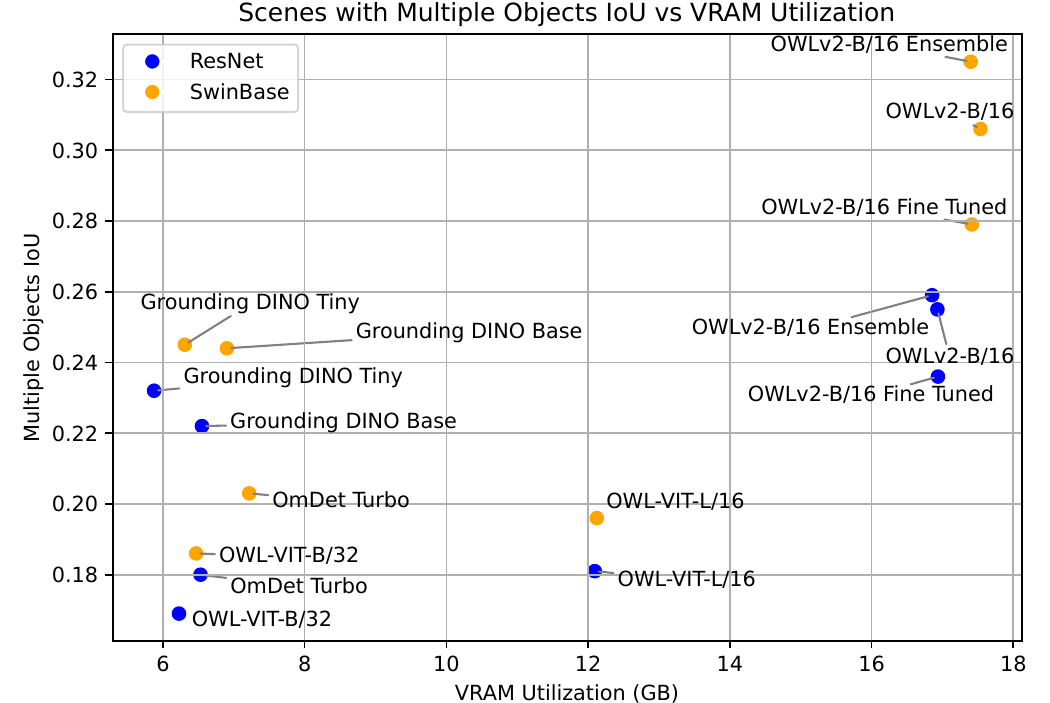}
    \includegraphics[width=0.48\linewidth]{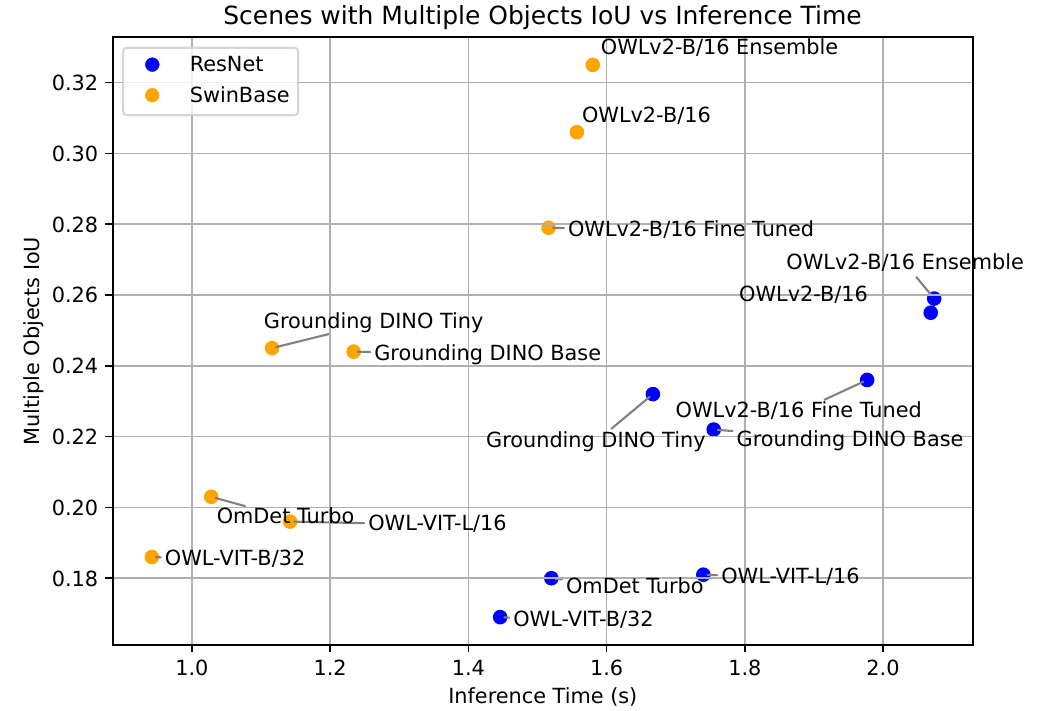}
    \caption{Accuracy-efficiency trade-offs (left: VRAM, right: inference time) for evaluated detector + part-segmentation combinations on AnyPart-Vision. SwinBase (orange) consistently outperforms ResNet-50 (blue) in both accuracy and speed when paired with the same detector. 
    OWLv2 detectors achieve the best IoU overall, but GroundingDino Tiny / Base with SwinBase provides a practical compromise between accuracy, VRAM requirements, and latency. }
    
    \label{fig:iou-vs-vram-time}
\end{figure*}

\textbf{3) Multiple combinations offer near real-time performance.}  
As shown in Table~\ref{tab:Vision-Results}, most configurations operate under 2 seconds per image, one (OWL-VIT-B/32) even below 1 second. Figure~\ref{fig:iou-vs-vram-time} visualises the trade-offs between segmentation quality, inference time, and VRAM usage.

These results partially support \textbf{Claim 1} and \textbf{Claim 2}: high-quality, part-specific segmentation is achievable with pretrained vision-language models in near real-time. To complete support for these claims, our second suite of experiments evaluates the full AnyPart pipeline, including perception and grasping.

\begin{figure}[t]
    \centering
    \includegraphics[width=0.8\linewidth]{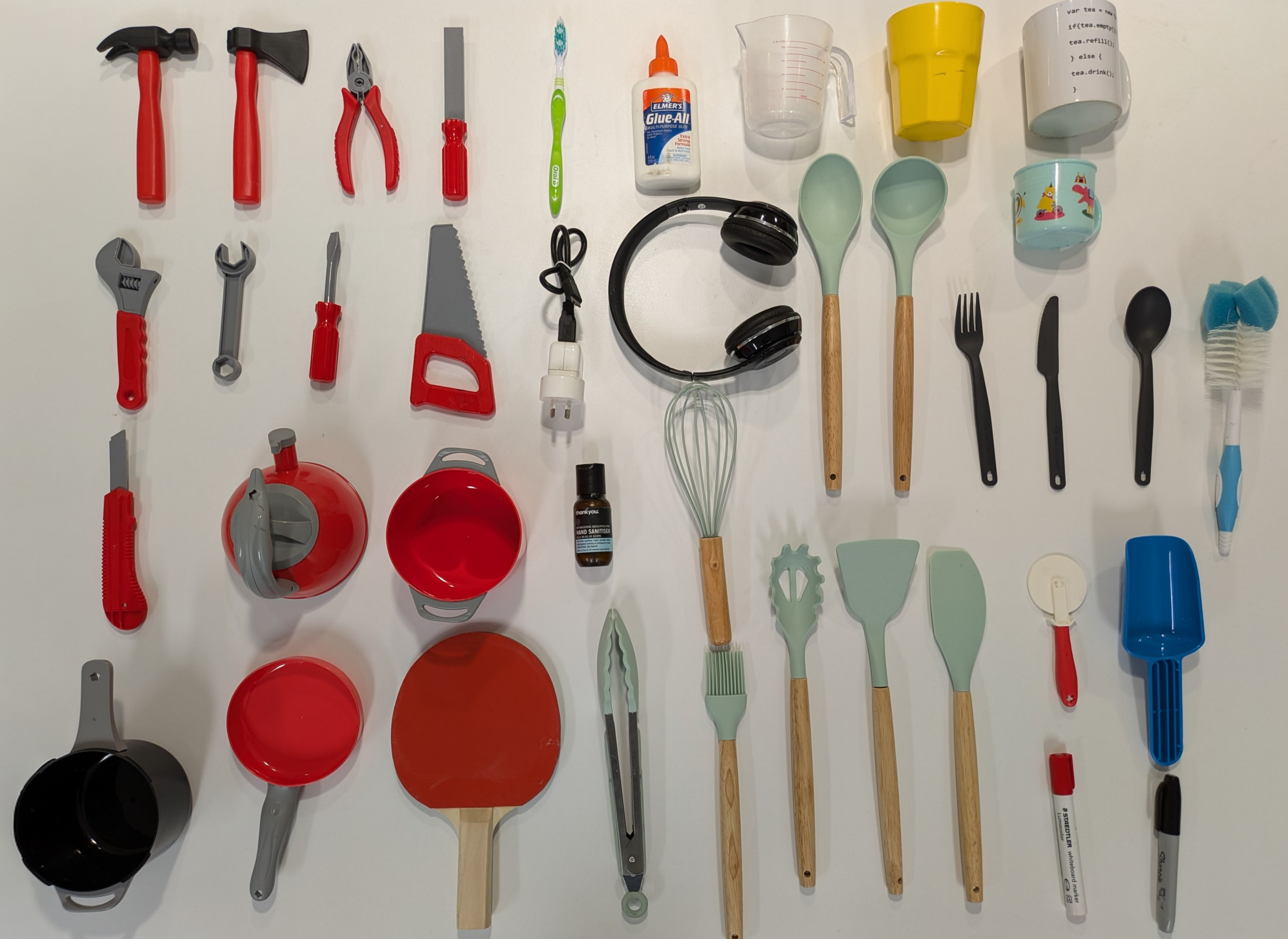}
    \caption{The items used for the real world grasping evaluations}
    \label{Physical-Grasping-Dataset}
\end{figure}

\subsection{Experiment 2: Real-World Grasping on AnyPart-Physical}

\begin{table*}[t]
  \centering
  \caption{%
  Full list of objects and associated parts used in the grasping experiments. The dataset includes 39 unique objects, 27 unique part types, and 76 total object-part combinations.} 
  \begin{tabular}{@{}lll|lll|lll|lll@{}} 
    \toprule
    \textbf{Object} & \textbf{Part 1} & \textbf{Part 2} &
    \textbf{Object} & \textbf{Part 1} & \textbf{Part 2} &
    \textbf{Object} & \textbf{Part 1} & \textbf{Part 2} &
    \textbf{Object} & \textbf{Part 1} & \textbf{Part 2} \\
    \midrule
    frying pan        & bowl     & handle   & pot              & rim      & handle   & dutch oven     & rim      & handles   & tea kettle       & handle   & spout    \\
    knife             & handle   & blade    & fork             & handle   & prongs   & spoon          & handle   & bowl      & hammer           & handle   & head     \\
    screwdriver       & handle   & shaft    & wrench           & handle   & head     & headphones      & earcup   & headband  & racket            & handle   & -        \\
    trowel            & scoop    & handle   & blue mug         & handle   & rim      & white mug       & handle   & rim       & file              & handle   & blade    \\
    exacto knife      & handle   & blade    & marker           & cap      & body     & sanitiser       & cap      & body      & scrub brush       & handle   & sponge   \\
    glue bottle       & cap      & body     & sharpie          & cap      & body     & charger         & body     & cable     & toothbrush        & handle   & bristles \\
    pliers            & handles  & gripper  & axe              & handle   & blade    & saw             & handle   & blade     & measuring cup     & handle   & rim      \\
    serving spoon     & bowl     & handle   & soup ladle       & bowl     & handle   & spatula         & flipper  & handle    & whisk             & whisk    & handle   \\
    food brush        & brush    & handle   & tongs            & grippers & handles  & pasta spoon      & strainer & handle    & food scraper       & scraper  & handle   \\
    spanner           & handle   & head     & pizza cutter     & handle   & slicer   & yellow cup       & rim      & -         &                   &          &          \\
    \bottomrule    
  \end{tabular}
  \label{tab:objects-parts}
\end{table*}

\begin{figure*}
    \centering
    \begin{subfigure}[b]{0.49\textwidth}
        \includegraphics[width=1\linewidth]{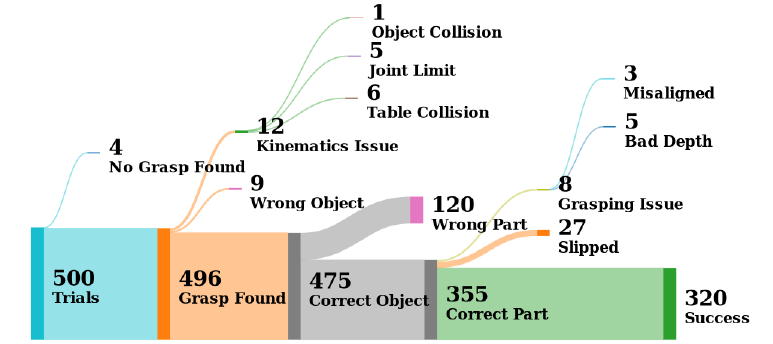}
        \caption{using Contact-GraspNet}
        \label{fig:sankey_contactgraspnet}
    \end{subfigure}
    \begin{subfigure}[b]{0.49\textwidth}
        \includegraphics[width=1\linewidth]{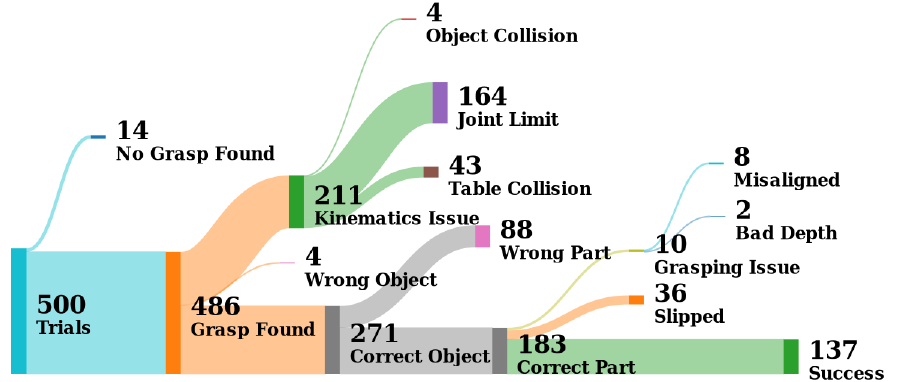}
        \caption{using AnyGrasp}
        \label{fig:sankey_anygrasp}
    \end{subfigure}\caption{%
    Sankey diagrams of outcome distributions for 500 real-world trials per grasp backend: \textbf{(a)} Contact-GraspNet and \textbf{(b)} AnyGrasp. 
    Contact-GraspNet shows a much larger success rate, whereas AnyGrasp routes more trials into kinematic/collision-related failures. Perception-induced errors (wrong part, occasional wrong object) appear in both. 
    The diagrams visualise how failure types differ by backend and support that many errors originate upstream in perception rather than grasp execution.}
  
    \label{fig:sankey}
\end{figure*}

\begin{figure}[t]
    \centering
    \includegraphics[width=1\linewidth]{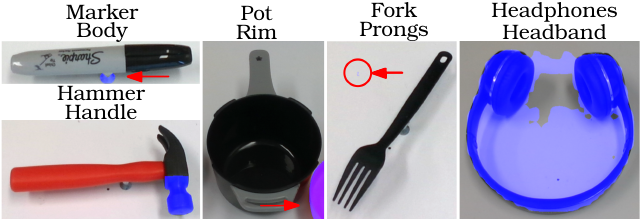}
     \caption{Failure cases induced by VLPart returning incomplete segmentations, segmentation of background elements, or over-segmentation. The prompt (object and part) is shown above the individual images.}

    \label{vlpart_failures}
\end{figure}

We evaluate AnyPart in 1,000 real-world grasping trials using a Franka Emika Panda arm equipped with a parallel-jaw gripper and a wrist-mounted Intel RealSense D435 RGB-D camera. The test set consists of 39 unique objects with 76 annotated part targets (see Table~\ref{tab:objects-parts} and Figure~\ref{Physical-Grasping-Dataset}). Each trial follows a four-step protocol:

\begin{enumerate}
    \item A natural language prompt specifying an object and part (\textit{e.g.}, ``mug, handle'') is issued.
    \item The robot captures an RGB-D image from the wrist-mounted camera.
    \item The image and prompt are passed through AnyPart, which predicts a 6-DoF grasp pose targeting the specified part.
    \item The robot executes the grasp using a velocity-controlled motion sequence, lifts the object 10\,cm above the table, and attempts to maintain a stable grasp.
\end{enumerate}

We conduct two types of experiments:

\begin{itemize}
    \item \textbf{Single-object scenes (760 trials):} Each object is placed individually on the table and rotated through 5 orientations (-90° to +90°, in 45° increments) relative to the robot. The target part is visible in all orientations.
    \item \textbf{Multi-object scenes (240 trials):} Several objects are placed in a cluttered tabletop scene with randomized positions and orientations.
\end{itemize}

All trials use the same hardware setup and fixed lighting. We intentionally do not perform grasp filtering or motion planning beyond direct execution of the predicted pose. This ensures we are evaluating the raw model predictions without bias from external heuristics or controllers.

A grasp is considered successful if the robot lifts the correct object by the correct part at least 10\,cm above the table in a stable fashion. The complete dataset of images, predictions, and outcomes is provided as the \textit{AnyPart-Physical} benchmark.

\textbf{Findings: } 
\textbf{1) Contact-GraspNet leads to substantially higher success rates.}  
Using Contact-GraspNet, AnyPart achieves a 65.0\% success rate in single-object scenes and 60.8\% in clutter. By contrast, AnyGrasp achieves only 29.7\% and 20.3\%, respectively (see Sankey diagrams in Fig.~\ref{fig:sankey}).

\textbf{2) Most failures originate from perception, not grasping.}  
Out of 1,000 trials, 208 failures were caused by inaccurate segmentation (under- or over-segmentation), while an additional 13 involved the wrong object being grasped due to poor bounding boxes. These trends support \textbf{Claim 3}, highlighting perception as the dominant bottleneck. Examples of these failures are shown in Figure~\ref{vlpart_failures}.

\textbf{3) AnyPart runs in real time.}  
Across all configurations, end-to-end inference time ranged between 0.9 and 2.1 seconds — over 60$\times$ faster than Lerf-TOGO’s 118 seconds. This supports \textbf{Claim 2} and demonstrates the practicality of our approach for real-time deployment.

\textbf{4) Modular foundation models generalise well out-of-the-box.}  
Despite zero retraining, AnyPart performs competitively with Lerf-TOGO~\cite{rashid2023language} and achieved 60.8\% vs.\ 69.0\% grasp success in muli-object scenes. These results affirm \textbf{Claim 1}. See Table~\ref{tab:lerf} for a detailed comparison of outcomes.
Although the experiments of Lerf-TOGO cannot be exactly replicated due to the set of objects, manipulator and environments being inaccessible to us, we designed our experiments to mirror the evaluation protocol used by Lerf-TOGO as best we could to enable a fair, if somewhat qualitative, comparison.

\section{Conclusion and Future Work}

This paper presented \textbf{AnyPart}, a modular and efficient framework for open-vocabulary part-based grasping that composes existing vision-language and grasp prediction models without retraining. Through a structured evaluation of 16 detector–segmenter combinations and 1,000 real-world grasping trials, we demonstrated that AnyPart achieves strong grasping performance while operating 60$\times$–100$\times$ faster than prior work like Lerf-TOGO.

Our results support three key claims: (1) pretrained foundation models can be effectively composed for part-based grasping; (2) the full pipeline runs in real time (1-2 seconds); and (3) perception—particularly open-vocabulary part segmentation—is the dominant bottleneck for performance.

We contribute two datasets to support further research: \textit{AnyPart-Vision} for benchmarking perception modules, and \textit{AnyPart-Physical}, which captures all observations, segmentations, and grasp outcomes from the real-world experiments.

\textbf{Future work} will explore using vision-language models not only for perception but also for action reasoning. We aim to extend AnyPart beyond grasp generation to open-vocabulary object manipulation, where the grasp pose is conditioned on intended downstream actions. Additionally, we will explore integrating image-to-text models to generate natural prompts autonomously, removing the need for hand-specified object-part queries.

We hope AnyPart provides a strong foundation for future research in modular, open-world robot manipulation, and encourages further benchmarking of vision-language models under real-world robotic constraints.

\begin{table}[tb]
  \centering
  \caption{Comparing AnyPart (using Contact-GraspNet) with  Lerf-TOGO~\cite{rashid2023language}.
 While Lerf-TOGO was tested on a different set of objects and manipulator, making it impossible to exactly replicate the experiments, we notice that AnyPart achieves comparable performance.}
  \label{tab:lerf}
  \begin{tabular}{@{}llll@{}} 
    \toprule
    \textbf{Outcome} & \multicolumn{2}{c}{\textbf{AnyPart (ours)}} & \textbf{Lerf-TOGO} \\
    \cmidrule(lr){2-3}
    & Single Object & Multi Object & \\   
    \midrule
    Success & 65.0\% & 60.8\% & 69.00\%\\
    Couldn't Produce Grasp & 0.8\% & 0.8\%  & - \\
    Grasp Depth issue & 1.3\% & 0.0\%  & - \\
    Grippers Slipped & 5.0\% & 6.7\% & -\\
    Grasped Wrong Part & 24.7\% & 21.7\% & 14\% \\
    Grasped Wrong Object & - & 7.5\% & 4\% \\
    Grasp Missaligned & 0.8\% & 0.0\% & - \\
    Kinematics Issue & 2.4\% & 2.5\% & - \\
    Unknown Failure & - & - & 13\% \\
    \bottomrule
  \end{tabular}
\end{table}

\bibliographystyle{IEEEtran}
\bibliography{refs}

\end{document}